\documentclass{article}

\usepackage[final]{neurips_2025}
\usepackage{natbib}
\usepackage{amssymb}

\usepackage[ruled,vlined,linesnumbered]{algorithm2e}

\usepackage{amsmath}
\usepackage{graphicx}
\usepackage{booktabs}
\usepackage{hyperref}

\usepackage{hyperref}

\title{
Focus Is All You Need:
Adaptive Goal-aware Attention Orchestration for Multi-Agent Graph Systems
}

\author{
  Mingzhou Fan \\
  Siyuan Xu \\
  Mingxuan Yuan
}

\begin{document}
\makeatletter
\providecommand{\@trackname}{}
\makeatother

\maketitle

% ------------------------------------------------
% Abstract
% ------------------------------------------------

\begin{abstract}

Large Language Models (LLMs) have enabled increasingly powerful autonomous
agents capable of reasoning, planning, and tool utilization. Recent advances
have shifted agent systems from single-agent execution loops toward
multi-agent graph-based workflows, where complex objectives are decomposed
into interconnected specialized agents. While graph-based orchestration
provides flexible task organization and coordination, it introduces a new
fundamental challenge: \textbf{attention allocation}. As the number of agents
and workflow nodes increases, existing systems often execute large portions
of the graph uniformly, causing computational resources to be distributed
across irrelevant or low-impact tasks.

In this work, we argue that future agentic systems require not only reasoning
capability but also the ability to dynamically allocate computational focus.
Inspired by the attention mechanism in Transformer architectures, we
introduce \textbf{Attention Orchestration}, a new paradigm that extends
attention from token-level representation learning to workflow-level agent
coordination. Specifically, we propose
\textbf{Adaptive Goal-aware Attention Orchestration (AGAO)}, a framework that
dynamically estimates agent importance according to user objectives, graph
dependencies, and computational constraints.

AGAO integrates three complementary components: (1) goal-aware attention,
which measures semantic relevance between user goals and agent capabilities;
(2) topology-aware attention, which incorporates structural dependencies
within agent graphs; and (3) resource-aware attention, which adaptively
allocates computational budgets and execution priorities among heterogeneous
agents. By introducing attention into the orchestration layer, AGAO transforms
static agent graphs into adaptive execution systems capable of focusing on
goal-critical reasoning paths.

Extensive experiments across diverse multi-agent workloads demonstrate that
AGAO improves task effectiveness while reducing unnecessary computation,
latency, and token consumption compared with existing graph-based agent
execution strategies. Our work establishes \textbf{Attention Engineering} as
a new direction for building scalable and intelligent multi-agent graph
systems. Code is available at \url{https://github.com/MingzhouFan97/AGAO}.

\end{abstract}

% ------------------------------------------------
% Main Paper
% ------------------------------------------------

\section{Introduction}

Large Language Models (LLMs) have demonstrated remarkable capabilities in
reasoning, planning, and tool utilization, leading to the emergence of
autonomous agent systems. Early LLM-based agents primarily relied on
iterative execution loops, where a single agent repeatedly performed
reasoning, action, and observation until completing a predefined objective.
Although effective for relatively simple tasks, such loop-based agents
struggle with increasingly complex problems that require heterogeneous
capabilities, parallel exploration, and collaborative reasoning.

\begin{figure}[htbp] % htbp 是位置参数，表示依次尝试放在此处、页顶、页底、单独一页[reference:10]
    \centering % 使图片居中[reference:11]
    \includegraphics[trim={50bp 185bp 20bp 90bp}, clip, width=1.02\textwidth]{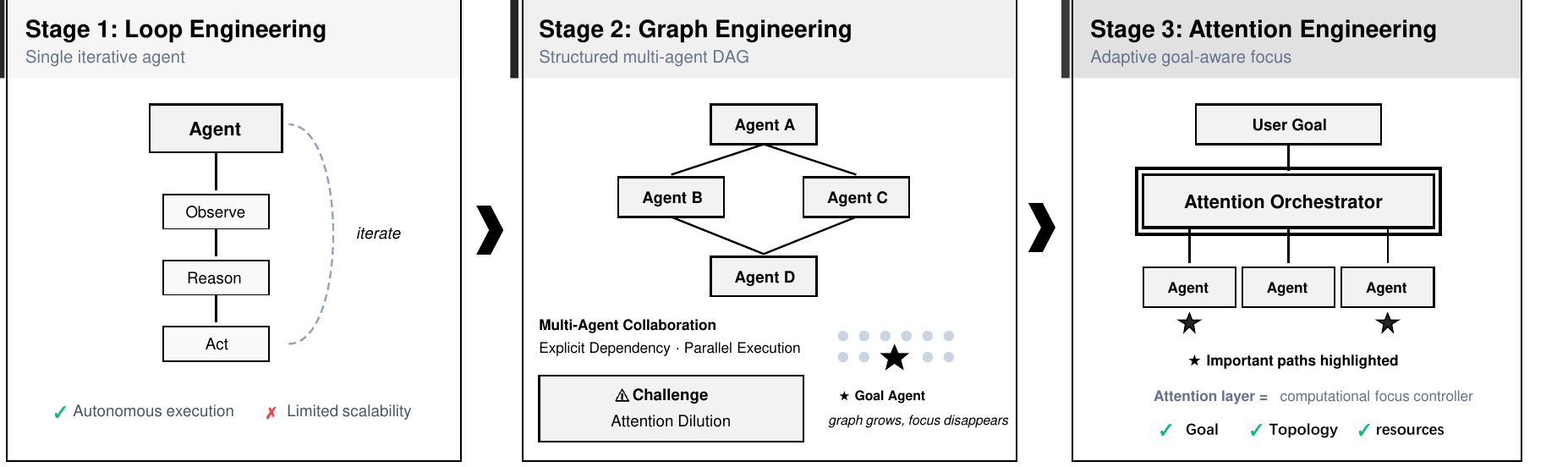}
    \caption{Evolution of autonomous agent systems.
Loop Engineering enables iterative single-agent reasoning,
Graph Engineering enables structured multi-agent collaboration,
while Attention Engineering introduces adaptive computational focus over
large-scale agent graphs.} % 添加带编号的标题[reference:12]
    \label{fig:my_label} % 添加标签，方便在文中用 \ref{fig:my_label} 引用[reference:13]
\end{figure}

To address these limitations, recent research has shifted from
\textit{agent loops} toward \textit{multi-agent graph systems}. By
representing complex tasks as directed graphs, Graph-based agent frameworks
enable explicit task decomposition, dependency modeling, and flexible
coordination among specialized agents. Modern agent orchestration systems,
such as workflow-based and graph-based frameworks, have demonstrated the
ability to scale agent collaboration beyond the capability of a single
reasoning loop.

However, while graph-based orchestration provides agents with structural
organization, it introduces a fundamental challenge that remains largely
unexplored: \textbf{computational focus allocation}. Existing multi-agent
graph systems typically assume that all executable nodes contribute equally
to the final objective. Consequently, when the number of agents increases,
the system may suffer from attention dilution, where substantial computational
resources are allocated to irrelevant or low-impact tasks. In other words,
current graph-based agents can determine \textit{which tasks exist and how
they are connected}, but lack the ability to determine
\textit{which tasks deserve attention at a given moment}.

For example, a large-scale research workflow may contain dozens or hundreds
of specialized agents responsible for information retrieval, analysis,
verification, and synthesis. However, the importance of each agent is highly
dependent on the evolving user objective. A market-analysis agent may become
critical when investigating business competition, while a technical-analysis
agent may dominate when evaluating engineering feasibility. Static graph
execution strategies fail to capture such dynamic changes in task relevance,
leading to unnecessary computation, increased latency, and degraded reasoning
quality.

Inspired by the success of attention mechanisms in Transformer architectures,
we argue that attention should not be limited to token-level representation
learning, but should be extended to the orchestration layer of agent systems.
In Transformer models, attention enables selective information aggregation by
dynamically assigning importance among input tokens. Analogously, future
agentic systems require mechanisms that dynamically allocate computational
focus among agents, tasks, and execution paths.

In this work, we introduce \textbf{Attention Orchestration}, a new paradigm
for adaptive focus allocation in multi-agent graph systems. Specifically, we
propose \textbf{Adaptive Goal-aware Attention Orchestration (AGAO)}, which
extends the concept of attention from neural architectures to workflow-level
agent coordination. Given a user goal, AGAO dynamically estimates the
relevance of each agent, incorporates graph structural dependencies, and
allocates computational resources according to evolving task importance.

Our framework consists of three complementary attention mechanisms:
(1) \textit{Goal-aware Attention}, which measures the semantic relevance
between user objectives and agent capabilities;
(2) \textit{Topology-aware Attention}, which incorporates graph dependencies
and execution structure into attention estimation; and
(3) \textit{Resource-aware Attention}, which dynamically assigns model
capacity, execution priority, and computational budgets according to
attention scores.

By introducing attention into the orchestration layer, we transform static
agent graphs into adaptive computational systems that can selectively focus
on goal-critical reasoning paths. We demonstrate that Attention Orchestration
can improve task effectiveness while reducing unnecessary computation across
diverse multi-agent workloads.

The main contributions of this work are summarized as follows:

\begin{itemize}

\item We introduce \textbf{Attention Engineering}, a new paradigm that extends
attention mechanisms from neural representation learning to multi-agent
workflow orchestration, enabling adaptive computational focus allocation.

\item We propose \textbf{Adaptive Goal-aware Attention Orchestration (AGAO)},
a hierarchical attention framework that integrates goal relevance, graph
topology, and resource allocation for dynamic multi-agent graph execution.

\item We develop an attention-driven agent execution framework and conduct
comprehensive experiments on diverse multi-agent workloads, demonstrating
improvements in task performance, computational efficiency, and scalability.

\end{itemize}

\section{Related Work}

\subsection{LLM-based Autonomous Agents}

Recent advances in Large Language Models (LLMs) have enabled autonomous
agents capable of reasoning, planning, and interacting with external tools.
ReAct~\cite{yao2023react} introduced an interleaved reasoning and acting
paradigm, allowing language models to iteratively reason about goals and
execute external actions. Toolformer~\cite{schick2023toolformer} further
explored self-supervised tool usage, demonstrating the potential of LLMs to
augment their capabilities through external APIs.

Beyond individual agent loops, autonomous agent frameworks such as
AutoGPT~\cite{autogpt2023} and AutoGen~\cite{wu2024autogen} explored more
general task execution paradigms through planning, tool invocation, and
multi-agent interaction. Recent systems such as GPT-Engineer~\cite{gptengineer2023}
and SWE-agent~\cite{yang2024sweagent} further extended autonomous agents
towards software engineering scenarios by integrating planning, code
generation, and environment feedback.

However, existing autonomous agent frameworks usually rely on predefined
execution strategies or manually specified interaction patterns. The ability
to dynamically allocate computational focus among heterogeneous agents during
execution remains largely unexplored.

\subsection{Multi-Agent Collaboration and Workflow Optimization}

Multi-agent LLM systems aim to solve complex tasks through collaboration
among specialized agents. CAMEL~\cite{camel2023} introduced role-playing
agents to study scalable communication among LLM-based agents.
MetaGPT~\cite{hong2023metagpt} and ChatDev~\cite{qian2024chatdev}
demonstrated structured agent collaboration for software engineering tasks.

AutoGen~\cite{wu2024autogen} further provided a flexible framework where
multiple customizable agents communicate through programmable interaction
patterns. Recent approaches such as DyLAN~\cite{liu2024dylan} dynamically
select LLM agents and adapt team configurations based on execution feedback.

Beyond agent collaboration, several works investigate automatic workflow
construction and planning. Plan-and-Solve prompting~\cite{wang2023plan}
introduced explicit planning stages before execution, while AFlow~\cite{zhang2025aflow}
formulated agentic workflow generation as a search problem over candidate
execution graphs.

Although these approaches significantly improve agent specialization and
workflow construction, most systems optimize agent selection or workflow
generation before execution. They generally lack a runtime mechanism that
continuously determines which agents, paths, and resources should receive
attention according to evolving goals and intermediate states.

\subsection{Graph-based Agent Orchestration}

Graph structures have recently emerged as an effective abstraction for
representing complex reasoning processes and agent workflows.
Graph of Thoughts~\cite{besta2023graph} explored graph-based reasoning by
organizing intermediate reasoning states as structured computation graphs.

For practical agent systems, LangGraph~\cite{langgraph2024} introduced a
graph-based framework for building stateful multi-agent applications, where
agents, tools, and states are represented as nodes connected through explicit
execution edges. Such graph-based abstractions improve controllability,
debuggability, and modularity.

However, existing graph-based agent frameworks mainly focus on representing
and executing predefined workflows. The graph topology, agent activation
patterns, and communication structures are typically specified manually or
optimized offline. Even workflow search approaches such as AFlow~\cite{zhang2025aflow}
primarily optimize graph construction rather than runtime execution dynamics.

In contrast, our work introduces an adaptive attention layer over agent
graphs, enabling runtime prioritization of agents, graph paths, and
computational resources according to task goals and execution feedback.

\subsection{Attention Mechanisms and Dynamic Routing}

Attention mechanisms have become a fundamental paradigm for selective
information processing. Transformer~\cite{vaswani2017attention} introduced
self-attention for dynamically modeling relationships among tokens.

Beyond language modeling, Graph Attention Networks (GAT)~\cite{velickovic2018gat}
extended attention mechanisms to graph structures by learning adaptive
importance weights among neighboring nodes.

Mixture-of-Experts (MoE) architectures further explored dynamic routing by
selecting specialized expert networks for different inputs
\cite{shazeer2017outrageously,fedus2022switch}. Recently, Mixture-of-Agents
(MoA)~\cite{wang2024mixture} extended the MoE paradigm to large-scale agent
collaboration.

However, existing attention and routing mechanisms mainly operate within
neural architectures, optimizing token representations, node embeddings, or
model computation. They do not address execution-level attention in
multi-agent systems, where routing decisions must consider semantic goals,
workflow topology, heterogeneous agent capabilities, and external resources.

\subsection{Positioning of Our Work}

Different from previous studies, we introduce
\textbf{Attention Orchestration} as a new abstraction layer for adaptive
agentic systems.

Unlike Transformer attention, which operates over token representations,
our attention mechanism operates over executable agents and workflow
structures.

Unlike Graph Attention Networks, which optimize node representations, our
approach optimizes graph execution decisions.

Unlike Mixture-of-Experts routing, which selects neural experts or model
components, our framework routes heterogeneous agents, tools, and
computational resources.

Therefore, AGAO explores a new research direction:
\textbf{goal-aware attention allocation for adaptive multi-agent graph
systems}, where attention becomes an execution-level control mechanism rather
than merely a representation-level operation.

Table~\ref{tab:paradigm_comparison} summarizes the conceptual differences
between existing attention mechanisms, graph-based approaches, and our
proposed attention orchestration paradigm. Unlike previous methods that
optimize representations, embeddings, or workflow structures, AGAO performs
execution-level attention allocation over heterogeneous agents, graph paths,
and computational resources.

\begin{table}[htb]
\centering
\fontsize{8.5pt}{10.2pt}\selectfont
\caption{Comparison of attention paradigms across different computational
levels.}
\label{tab:paradigm_comparison}
\begin{tabular}{lccc}
\toprule
\textbf{Paradigm} &
\textbf{Attention Target} &
\textbf{Optimization Focus} &
\textbf{Dynamic Routing} \\
\midrule

Transformer &
Tokens &
Representation &
\checkmark \\

Graph Attention Network &
Nodes &
Embedding &
\checkmark \\

Mixture-of-Experts &
Experts &
Computation &
\checkmark \\

Multi-Agent Workflow &
Agents &
Workflow &
$\times$ \\

\textbf{AGAO (Ours)} &
\textbf{Agents + Graph + Resources} &
\textbf{Goal-aware Orchestration} &
\checkmark \\

\bottomrule
\end{tabular}
\end{table}

\section{Problem Formulation}

\subsection{Multi-Agent Graph Systems}

We consider a multi-agent graph system represented as a directed graph:

\begin{equation}
G=(V,E)
\end{equation}

where $V=\{v_1,v_2,...,v_N\}$ denotes a set of agent nodes and
$E$ represents directed dependencies among agents.

Each agent node $v_i$ is associated with a heterogeneous capability profile:

\begin{equation}
v_i=(x_i,m_i,c_i)
\end{equation}

where $x_i$ represents the semantic capability description of the agent,
$m_i$ denotes the underlying model or execution engine, and $c_i$ represents
the computational cost of invoking the agent.

Unlike traditional computational graphs where nodes usually perform fixed
operations, agent nodes are autonomous decision-making units that may
generate different outputs depending on task objectives and intermediate
states.

Given an input objective $q$, the graph system aims to execute a subset of
agents and aggregate their outputs to generate a final response:

\begin{equation}
y = F(G,q)
\end{equation}

where $F(\cdot)$ represents the graph execution policy.

\subsection{Goal-aware Agent Relevance}

In existing graph-based agent systems, execution decisions are primarily
determined by predefined graph structures. All reachable nodes are often
treated with similar priority regardless of their relevance to the current
objective.

However, in real-world tasks, the importance of each agent is dynamically
dependent on the user goal.

We define the relevance between a user objective $q$ and an agent node $v_i$
as:

\begin{equation}
r_i = R(q,v_i)
\end{equation}

where $R(\cdot)$ measures how much the capability of agent $v_i$ contributes
to achieving objective $q$.

The ideal execution policy should prioritize agents with high relevance while
suppressing unnecessary computation from low-impact agents.

\subsection{Attention Allocation Problem}

Inspired by Transformer attention mechanisms, we formulate agent selection as
an attention allocation problem.

Given a goal representation:

\begin{equation}
Q_g \in \mathbb{R}^{d}
\end{equation}

and agent representations:

\begin{equation}
K_A=\{k_1,k_2,...,k_N\}
\end{equation}

the attention score of each agent is defined as:

\begin{equation}
\alpha_i=
\frac{
\exp(Q_g k_i^T/\sqrt{d})
}
{
\sum_{j=1}^{N}
\exp(Q_g k_j^T/\sqrt{d})
}
\end{equation}

where $\alpha_i$ indicates the computational focus assigned to agent
$v_i$.

The attention distribution satisfies:

\begin{equation}
\sum_{i=1}^{N}\alpha_i=1
\end{equation}

A higher attention score indicates that the corresponding agent is more
critical for accomplishing the current objective.

\subsection{Graph-aware Execution Objective}

Although semantic relevance provides an important signal, agent execution
decisions cannot rely solely on independent relevance scores. Agents are
connected through dependencies, and some intermediate nodes may become
important due to their structural position in the graph.

Therefore, we define the overall attention score as:

\begin{equation}
\alpha_i =
f(\alpha_i^{goal},
\alpha_i^{topo},
\alpha_i^{resource})
\end{equation}

where:

\begin{itemize}
    \item $\alpha_i^{goal}$ represents semantic relevance between the goal
    and the agent;
    \item $\alpha_i^{topo}$ captures graph structural importance;
    \item $\alpha_i^{resource}$ represents adaptive computation allocation.
\end{itemize}

\subsection{Optimization Objective}

The objective of adaptive attention orchestration is to maximize task utility
while minimizing unnecessary computation.

We formulate the optimization problem as:

\begin{equation}
\max_{\pi}
\mathcal{J}(\pi)
=
\mathcal{Q}(y,q)
-
\lambda_1 \mathcal{C}(\pi)
-
\lambda_2 \mathcal{L}(\pi)
\end{equation}

where:

\begin{itemize}

\item $\pi$ denotes the agent execution policy;

\item $\mathcal{Q}(y,q)$ measures task completion quality;

\item $\mathcal{C}(\pi)$ represents computational cost, including token
consumption and model invocation cost;

\item $\mathcal{L}(\pi)$ represents execution latency;

\item $\lambda_1$ and $\lambda_2$ are trade-off coefficients.

\end{itemize}

The goal of our framework is therefore to learn an adaptive execution policy
that dynamically allocates attention across agents while preserving task
quality under limited computational budgets.

\section{Adaptive Goal-aware Attention Orchestration}

\subsection{Framework Overview}

Existing multi-agent graph systems typically execute workflows according to
predefined graph structures, where task dependencies are explicitly specified
but computational focus remains static. In contrast, we propose
\textbf{Adaptive Goal-aware Attention Orchestration (AGAO)}, which introduces
an attention-driven execution layer between user objectives and agent graph
execution.

The key idea of AGAO is to treat agents as dynamically selectable
computational units rather than fixed workflow operators. Given an evolving
task objective, AGAO continuously estimates the importance of each agent and
adjusts execution priorities, routing decisions, and computational resources.

\begin{figure}[htbp] % htbp 是位置参数，表示依次尝试放在此处、页顶、页底、单独一页[reference:10]
    \centering % 使图片居中[reference:11]
    \includegraphics[trim={130bp 100bp 20bp 20bp}, clip, width=1.25\textwidth]{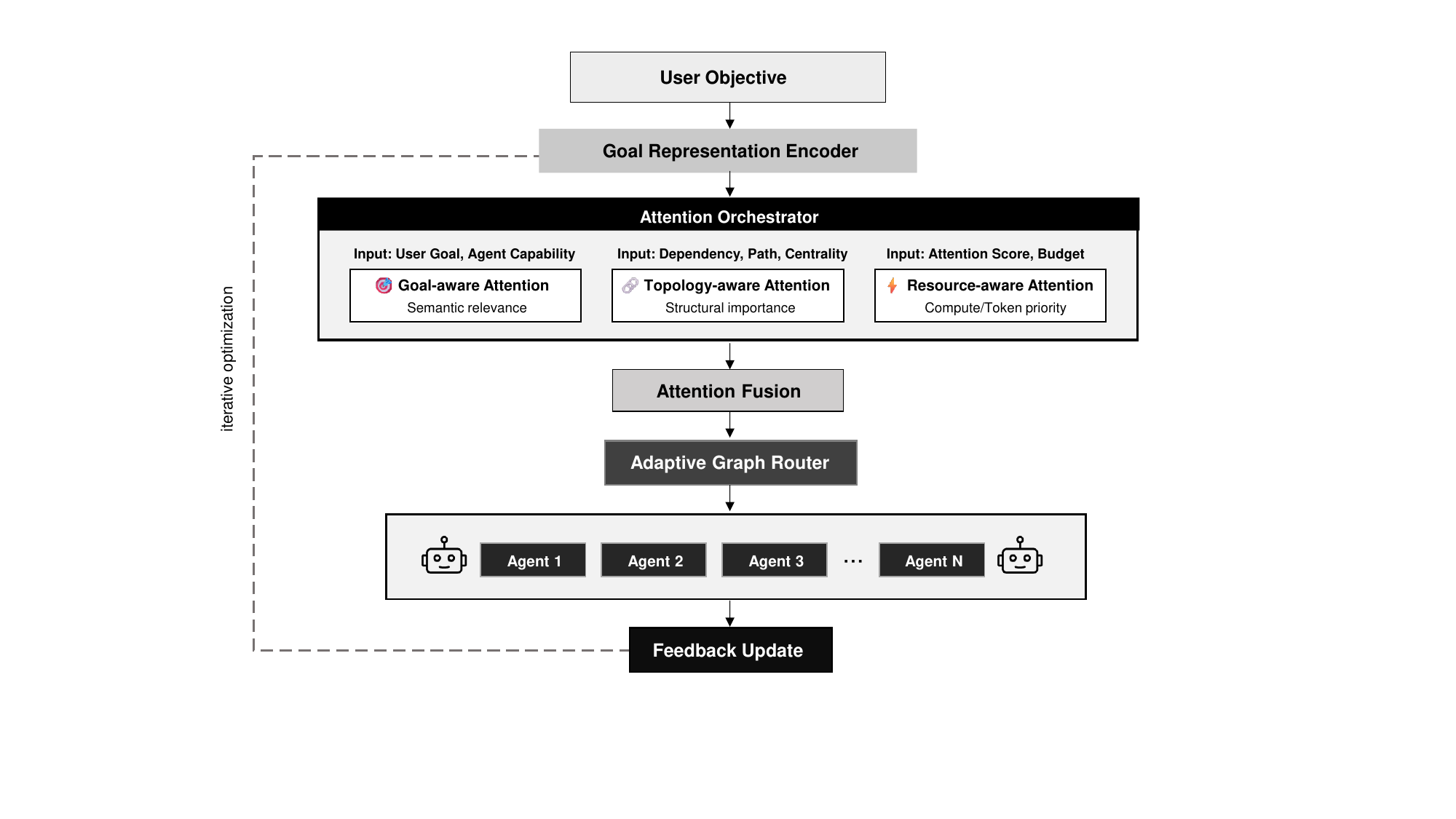}
    \caption{Overview of Adaptive Goal-aware Attention Orchestration (AGAO).
The framework introduces an attention layer between user objectives and
multi-agent graph execution, integrating semantic relevance, graph topology,
and computational resource constraints to dynamically control agent
activation and execution.} % 添加带编号的标题[reference:12]
    \label{fig:my_label} % 添加标签，方便在文中用 \ref{fig:my_label} 引用[reference:13]
\end{figure}

The overall architecture of AGAO is illustrated as follows:

\begin{equation}
\text{Goal}
\rightarrow
\text{Attention Orchestrator}
\rightarrow
\text{Adaptive Agent Graph}
\rightarrow
\text{Execution Feedback}
\end{equation}

Specifically, AGAO consists of three hierarchical attention modules:

\begin{enumerate}

\item \textbf{Goal-aware Attention}: 
captures the semantic alignment between the user objective and individual
agent capabilities.

\item \textbf{Topology-aware Attention}:
incorporates graph structural information, dependency relationships, and
critical execution paths into attention estimation.

\item \textbf{Resource-aware Attention}:
translates attention scores into practical execution decisions, including
model selection, token budget allocation, and agent activation priority.

\end{enumerate}

Unlike traditional graph execution strategies that apply a fixed computation
policy, AGAO produces an adaptive attention distribution:

\begin{equation}
\alpha_t=
\{\alpha_1^t,\alpha_2^t,...,\alpha_N^t\}
\end{equation}

where $\alpha_i^t$ represents the computational focus assigned to agent
$v_i$ at execution step $t$.

As new information is generated during execution, AGAO updates attention
allocation based on intermediate observations:

\begin{equation}
\alpha_{t+1}
=
\Phi(\alpha_t,o_t,G,q)
\end{equation}

where $o_t$ denotes intermediate observations and $\Phi(\cdot)$ represents
the adaptive attention update function.

Through this mechanism, AGAO enables multi-agent graph systems to dynamically
focus on goal-critical reasoning paths while reducing unnecessary execution
overhead.

\subsection{Goal-aware Attention}

The first challenge in multi-agent graph execution is determining which
agents are relevant to the current objective. Unlike traditional workflow
systems where node importance is predefined, agent relevance is highly
dependent on the evolving user goal.

To address this problem, we introduce \textbf{Goal-aware Attention}, which
computes the semantic alignment between the current objective and each agent
node.

Given a user goal representation:

\begin{equation}
q \in \mathbb{R}^{d_q}
\end{equation}

we encode the goal into a query vector:

\begin{equation}
Q_g = W_q f_g(q)
\end{equation}

where $f_g(\cdot)$ represents a goal encoder and $W_q$ is a learnable
projection matrix.

For each agent node $v_i$, we construct an agent representation containing
its capability description, historical execution information, and current
task state:

\begin{equation}
z_i=
[x_i;h_i;s_i]
\end{equation}

where:

\begin{itemize}
    \item $x_i$ denotes the semantic capability embedding of agent $v_i$;
    \item $h_i$ represents historical execution statistics;
    \item $s_i$ represents the current execution state.
\end{itemize}

The agent representation is projected into key and value vectors:

\begin{equation}
K_i=W_k z_i
\end{equation}

\begin{equation}
V_i=W_v z_i
\end{equation}

The goal-aware attention score is computed as:

\begin{equation}
e_i=
\frac{
Q_g K_i^T
}
{\sqrt{d}}
\end{equation}

The normalized attention distribution over all agents is obtained by:

\begin{equation}
\alpha_i^{goal}
=
\frac{
\exp(e_i)
}
{
\sum_{j=1}^{N}
\exp(e_j)
}
\end{equation}

where $\alpha_i^{goal}$ indicates the semantic importance of agent $v_i$
with respect to the current goal.

The attended agent representation is then computed as:

\begin{equation}
C_g=
\sum_{i=1}^{N}
\alpha_i^{goal}V_i
\end{equation}

Unlike static workflow execution, where all reachable agents receive equal
execution priority, Goal-aware Attention dynamically concentrates computation
on agents that contribute most to the current objective.

\subsection{Topology-aware Attention}

Although Goal-aware Attention captures semantic relevance between user
objectives and agent capabilities, agent execution in graph-based systems is
not independent. The importance of an agent is also determined by its
structural position, dependency relationships, and influence on downstream
execution.

Therefore, we introduce \textbf{Topology-aware Attention}, which incorporates
graph structural information into attention estimation.

Given an agent graph:

\begin{equation}
G=(V,E)
\end{equation}

we define a topology representation for each agent node $v_i$:

\begin{equation}
p_i=f_{topo}(G,v_i)
\end{equation}

where $f_{topo}(\cdot)$ extracts structural features including dependency
distance, predecessor relationships, successor influence, and graph centrality.

The topology representation is projected into a graph bias term:

\begin{equation}
B_{ij}=g(p_i,p_j)
\end{equation}

where $B_{ij}$ measures the structural relationship between agent nodes
$v_i$ and $v_j$.

Different from conventional self-attention:

\begin{equation}
A=softmax(\frac{QK^T}{\sqrt{d}})
\end{equation}

we introduce graph-aware attention:

\begin{equation}
A^{topo}
=
softmax(
\frac{QK^T}{\sqrt{d}}
+
B_G
)
\end{equation}

where $B_G$ represents the topology bias matrix derived from the agent graph.

The topology-aware importance of each agent is then computed as:

\begin{equation}
\alpha_i^{topo}
=
\sum_j A^{topo}_{ij}
\end{equation}

In addition to local dependency relationships, we further consider global
execution influence. Specifically, agents located on critical reasoning paths
should receive higher attention even when their semantic relevance is limited.

We define the structural influence score as:

\begin{equation}
s_i^{critical}
=
\eta_1 d_i^{in}
+
\eta_2 d_i^{out}
+
\eta_3 c_i^{path}
\end{equation}

where:

\begin{itemize}
    \item $d_i^{in}$ denotes incoming dependency importance;
    \item $d_i^{out}$ denotes downstream influence;
    \item $c_i^{path}$ represents critical path contribution.
\end{itemize}

Finally, the topology-aware attention is combined with semantic attention:

\begin{equation}
\alpha_i^{graph}
=
\gamma
\alpha_i^{goal}
+
(1-\gamma)
\alpha_i^{topo}
\end{equation}

where $\gamma$ controls the balance between semantic relevance and graph
structure.

Through topology-aware attention, AGAO avoids purely semantic pruning and
preserves structurally essential agents required for successful execution.

\subsection{Resource-aware Attention}

While Goal-aware and Topology-aware Attention determine the importance of
agents from semantic and structural perspectives, practical agent systems
must further decide how computational resources should be allocated.

Different agents have different computational requirements. Assigning equal
resources to all agents may result in inefficient execution, excessive token
consumption, and unnecessary latency.

Therefore, we introduce \textbf{Resource-aware Attention}, which transforms
attention scores into adaptive execution policies.

Given the combined attention score:

\begin{equation}
\alpha_i^{*}
=
f(
\alpha_i^{goal},
\alpha_i^{topo}
)
\end{equation}

we define a resource allocation function:

\begin{equation}
r_i
=
R(\alpha_i^{*},c_i,b)
\end{equation}

where:

\begin{itemize}

\item $r_i$ denotes the allocated computational resource for agent $v_i$;

\item $c_i$ represents the intrinsic computational cost of agent $v_i$;

\item $b$ represents the total available computation budget.

\end{itemize}

The resource allocation problem can be formulated as:

\begin{equation}
\max_{r_1,...,r_N}
\sum_i
\alpha_i^{*}U_i(r_i)
\end{equation}

subject to:

\begin{equation}
\sum_i r_i \leq B
\end{equation}

where $U_i(\cdot)$ represents the utility gained from allocating resources to
agent $v_i$, and $B$ denotes the global computation budget.

\begin{figure}[htbp] % htbp 是位置参数，表示依次尝试放在此处、页顶、页底、单独一页[reference:10]
    \centering % 使图片居中[reference:11]
    \includegraphics[trim={10bp 300bp 50bp 90bp}, clip, width=1\textwidth]{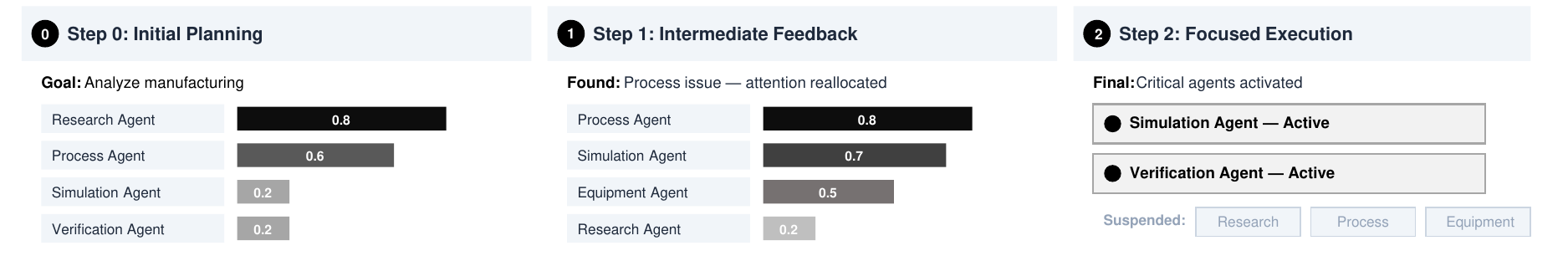}
    \caption{Adaptive attention evolution during multi-step agent execution.
AGAO dynamically reallocates attention according to intermediate feedback,
allowing the system to shift computational focus toward newly discovered
critical agents.} % 添加带编号的标题[reference:12]
    \label{fig:my_label} % 添加标签，方便在文中用 \ref{fig:my_label} 引用[reference:13]
\end{figure}

In practice, AGAO converts resource attention into three execution controls:

\paragraph{1. Model Routing}

Agents with high attention scores are assigned stronger reasoning models:

\begin{equation}
m_i=
\begin{cases}
M_{large}, & \alpha_i^{*}>\tau_h\\
M_{medium}, & \tau_l<\alpha_i^{*}\leq\tau_h\\
M_{small}, & \alpha_i^{*}\leq\tau_l
\end{cases}
\end{equation}

\paragraph{2. Token Budget Allocation}

The token budget is dynamically distributed:

\begin{equation}
T_i
=
T_{total}
\frac{\exp(\alpha_i^{*})}
{\sum_j\exp(\alpha_j^{*})}
\end{equation}

\paragraph{3. Execution Priority Scheduling}

The scheduler assigns execution priority according to attention:

\begin{equation}
P_i=\alpha_i^{*}\cdot w_i
\end{equation}

where $w_i$ represents execution urgency or dependency constraints.

Through Resource-aware Attention, AGAO transforms attention scores into
concrete execution decisions, enabling efficient scaling of large
multi-agent graphs under limited computational budgets.

\subsection{Adaptive Graph Routing}

Static attention allocation is insufficient for long-horizon agent tasks because
the importance of an agent may change during execution. An initially
irrelevant agent may become critical after observing intermediate results,
while previously important agents may become unnecessary.

To address this issue, we introduce \textbf{Adaptive Graph Routing}, where
attention distributions are dynamically updated according to execution
feedback.

At execution step $t$, the agent graph maintains an attention state:

\begin{equation}
\alpha_t=
\{\alpha_1^t,\alpha_2^t,...,\alpha_N^t\}
\end{equation}

After executing selected agents, the system receives intermediate
observations:

\begin{equation}
o_t=
\{y_t,s_t,e_t\}
\end{equation}

where:

\begin{itemize}
    \item $y_t$ denotes generated agent outputs;
    \item $s_t$ represents updated execution states;
    \item $e_t$ represents execution feedback signals.
\end{itemize}

The attention state is updated through an adaptive routing function:

\begin{equation}
\alpha_{t+1}
=
\Phi(
\alpha_t,
o_t,
G,
q
)
\end{equation}

where $\Phi(\cdot)$ represents the attention update policy.

Specifically, we decompose the updated attention into three components:

\begin{equation}
\alpha_{t+1}
=
\lambda_1
\alpha_{t}^{goal}
+
\lambda_2
\alpha_{t}^{topo}
+
\lambda_3
\alpha_{t}^{feedback}
\end{equation}

where:

\begin{itemize}

\item $\alpha_{t}^{goal}$ captures objective relevance;

\item $\alpha_{t}^{topo}$ preserves graph structural constraints;

\item $\alpha_{t}^{feedback}$ reflects execution-time observations.

\end{itemize}

The feedback attention is computed based on intermediate outcomes:

\begin{equation}
\alpha_i^{feedback}
=
g(y_i,
\Delta q,
r_i)
\end{equation}

where $\Delta q$ measures the remaining gap between the current execution
state and the original objective.

Based on updated attention scores, AGAO dynamically performs graph routing:

\begin{equation}
v_i^{t+1}
=
\begin{cases}
Execute,
&
\alpha_i^{t+1}>\tau_e
\\
Suspend,
&
\tau_s < \alpha_i^{t+1}\leq\tau_e
\\
Prune,
&
\alpha_i^{t+1}\leq\tau_s
\end{cases}
\end{equation}

Unlike conventional workflow engines that follow fixed execution paths,
adaptive graph routing enables agents to enter, leave, or change priority
during execution according to evolving task requirements.

\subsection{Execution Algorithm}

Based on the proposed attention mechanisms, we formulate AGAO as an adaptive
execution policy for multi-agent graph systems.

Given a task objective $q$, an agent graph $G=(V,E)$, and a computation budget
$B$, AGAO iteratively estimates agent importance, selects execution paths,
allocates resources, and updates attention according to intermediate feedback.

The complete procedure is summarized in Algorithm~\ref{alg:agao}.

\begin{algorithm}[t]
\caption{Adaptive Goal-aware Attention Orchestration (AGAO)}
\label{alg:agao}

\KwIn{
Goal $q$,
Agent Graph $G=(V,E)$,
Computation Budget $B$,
Thresholds $\tau_e,\tau_s$
}

\KwOut{
Final response $y$
}

Initialize execution state $S_0$ and attention distribution
$\alpha_0$\;

Encode user goal:
\[
Q_g=f_g(q)
\]

\While{task not completed}{

Compute goal-aware attention:

\[
\alpha^{goal}
=
Attention(Q_g,K_A,V_A)
\]

Compute topology-aware attention:

\[
\alpha^{topo}
=
Attention(Q_A,K_A,B_G)
\]

Fuse attention signals:

\[
\alpha_t
=
Fuse(
\alpha^{goal},
\alpha^{topo}
)
\]

Allocate computational resources:

\[
R_t
=
ResourceAllocator(\alpha_t,B)
\]

Select active agents:

\[
V_t
=
\{v_i|\alpha_i^t>\tau_e\}
\]

Execute selected agents with allocated resources:

\[
Y_t
=
Execute(V_t,R_t)
\]

Observe execution feedback:

\[
o_t=
Observe(Y_t,S_t)
\]

Update attention state:

\[
\alpha_{t+1}
=
\Phi(\alpha_t,o_t,G,q)
\]

Update graph state:

\[
S_{t+1}
=
Update(S_t,Y_t)
\]

}

Aggregate final outputs:

\[
y=Aggregate(Y_1,...,Y_T)
\]

\Return $y$

\end{algorithm}

\section{Experiments}
\subsection{Experimental Setting}

We evaluate the current AGAO implementation over 16 tasks: eight MBPP programming tasks \cite{austin2021program} and eight
HotpotQA multi-hop question-answering tasks \cite{yang2018hotpotqa}. Each task is executed with a
hand-specified nine-node graph. The coding graph contains roles for
specification analysis, implementation, testing, debugging, review, and
finalization; the question-answering graph contains decomposition,
retrieval, evidence selection, reasoning, verification, and finalization.
The task
manifest, graph templates, prompts, tools, retry behavior, and per-run
budget are fixed across policies.

We compare five execution policies. \textbf{Random} is a
dependency-valid, fixed-seed router that randomly selects up to two ready
nodes. \textbf{Static} executes every dependency-valid node according to
the fixed graph order.
\textbf{Semantic} selects up to two ready nodes according to lexical
overlap between the task prompt and the node capability description.
\textbf{MoE-style} selects up to two ready nodes using a fixed mixture of
lexical relevance and topology score, without feedback or adaptive
resource routing. \textbf{AGAO} combines lexical goal relevance, a
topology score, and rule-based feedback; it can skip low-scoring nodes
and routes selected nodes among low-, medium-, and high-capacity model
tiers.

For MBPP, we report pass@1 obtained by executing the supplied assertions.
For HotpotQA, we report token-level F1 after standard article and
punctuation normalization. We additionally report the number of active
nodes, the sum of recorded node execution times (\emph{cumulative agent
time}), and the input/output token counts returned by the gateway.
Cumulative agent time measures work consumed across calls; because calls
within a scheduling round can run concurrently, it is not end-to-end
wall-clock latency. Each policy is run once per task. The results are
therefore descriptive pilot means rather than estimates with statistical
significance.

\subsection{Experimental Results}

Table~\ref{tab:pilot-main} summarizes performance across the eight tasks in each evaluation suite. On MBPP, AGAO solves 7/8 tasks (pass@1 $=0.875$), compared with 6/8 for MoE-style routing, 5/8 for Static, and 4/8 for both Random and Semantic routing. These results demonstrate a coding advantage for AGAO on this benchmark.
On the QA evaluation set, AGAO achieves a mean token F1 of $0.340$, exceeding Static ($0.319$), Random ($0.285$), and Semantic routing ($0.244$), but falling below MoE-style routing ($0.417$). Taken together, the results suggest that AGAO’s effectiveness varies by workload, rather than consistently improving answer quality across all task types.

\begin{table*}[t]
\centering
% \small
\setlength{\tabcolsep}{4.5pt}
\caption{Each suite contains eight fixed tasks, and every
entry is a mean over one run per task. Cumulative agent time is the sum of
recorded per-node times, not end-to-end wall-clock latency.}
\label{tab:pilot-main}
\begin{tabular}{lrrrrrr}
\hline
& \multicolumn{3}{c}{MBPP} & \multicolumn{3}{c}{HotpotQA} \\
\cline{2-4}\cline{5-7}
Policy & pass@1 & nodes & time (s) & F1 & nodes & time (s) \\
\hline
Random    & 0.500 & 8.25 & 60.81 & 0.285 & 9.00 & 44.27 \\
Static    & 0.625 & 9.00 & 66.58 & 0.319 & 9.00 & 43.76 \\
Semantic  & 0.500 & 8.13 & 57.45 & 0.244 & 9.00 & 46.20 \\
MoE-style & 0.750 & 8.00 & 64.72 & \textbf{0.417} & 9.00 & 47.72 \\
\textbf{AGAO} & \textbf{0.875} & \textbf{6.50} & \textbf{29.33}
& 0.340 & \textbf{7.25} & \textbf{21.76} \\
\hline
\end{tabular}
\end{table*}

Across both suites, AGAO activates the fewest nodes and records the
lowest cumulative agent time. Relative to Static, it activates $27.8\%$
and $19.4\%$ fewer nodes on MBPP and HotpotQA, respectively, and reduces
cumulative agent time by $56.0\%$ and $50.3\%$. The same direction holds
relative to Random, Semantic, and MoE-style routing. These savings are the
strongest result of the pilot: AGAO reaches a better observed coding
score while executing fewer roles, and it maintains an intermediate QA
score with less agent work.

\subsection{Resource Accounting and Interpretation}

Total token count is not a direct measure of API cost because input and
output tokens are billed at different rates. Table~\ref{tab:pilot-tokens}
shows that AGAO uses more input tokens than the baselines, reflecting its
additional routing and context construction, but produces substantially
fewer output tokens. This trade-off is economically favorable because
output tokens are considerably more expensive than input tokens: under the
configured pricing, each output token costs six times as much as an input
token. Consequently, the reduction in generated tokens offsets the added
input context, making AGAO more cost-efficient even on relatively simple
tasks.

\begin{table*}[t]
\centering
% \small
\setlength{\tabcolsep}{3.5pt}
\caption{Mean gateway-reported token counts per task.}
\label{tab:pilot-tokens}
\begin{tabular}{lrrrrrr}
\hline
& \multicolumn{3}{c}{MBPP} & \multicolumn{3}{c}{HotpotQA} \\
\cline{2-4}\cline{5-7}
Policy & input & output & total & input & output & total \\
\hline
Random    & 1,087 & 2,869 & 3,955 & 2,258 & 2,085 & 4,343 \\
Static    & 1,339 & 3,233 & 4,572 & 2,304 & 2,069 & 4,373 \\
Semantic  & 1,176 & 2,755 & 3,931 & 2,304 & 2,098 & 4,402 \\
MoE-style & 1,150 & 2,684 & 3,835 & 1,866 & 2,223 & 4,089 \\
AGAO      & 9,670 & 1,390 & 11,059 & 5,573 & 1,168 & 6,741 \\
\hline
\end{tabular}
\end{table*}

\subsection{Ablation Study}

Table~\ref{tab:ablation} reports a unified ablation study over the same
eight tasks per suite. We remove one mechanism at a time from full AGAO:
the lexical goal score, topology score, or resource-routing mechanism.
We additionally report a semantic-only variant that removes topology,
resource routing, and feedback together. The table includes quality,
active-node count, and cumulative agent time so that the effects on both task outcomes and resource use are
visible.

\begin{table*}[t]
\centering
% \scriptsize
\setlength{\tabcolsep}{2.8pt}
\caption{Unified ablation study. All values are means over eight tasks per
suite and one run per task. w/o denotes removal of the named mechanism.
Time is cumulative agent time.}
\label{tab:ablation}
\begin{tabular}{lrrrrrr}
\hline
& \multicolumn{3}{c}{MBPP} & \multicolumn{3}{c}{HotpotQA} \\
\cline{2-4}\cline{5-7}
Variant & pass@1 & nodes & time (s) & F1 & nodes & time (s) \\
\hline
\textbf{AGAO} & 0.875 & 6.50 & 29.33 & 0.340 & 7.25 & \textbf{21.76} \\
AGAO w/o Goal & \textbf{1.000} & 5.88 & \textbf{23.17}
  & 0.295 & 7.00 & 43.72 \\
AGAO w/o Topology & 0.500 & 5.13 & 28.74
  & 0.266 & \textbf{6.25} & 33.58 \\
AGAO w/o Resource & 0.250 & 6.50 & 42.29
  & \textbf{0.370} & 7.25 & 35.68 \\
AGAO semantic-only & 0.375 & \textbf{4.88} & 31.73
  & 0.236 & 7.00 & 34.86 \\
\hline
\end{tabular}
\end{table*}

The ablation study highlights the value of combining AGAO’s three mechanisms—goal scoring, topology-aware prioritization, and resource-aware routing—to achieve stable performance across both evaluation suites. Although removing individual components can improve a single metric in isolation, these changes introduce clear trade-offs on the other workload. For example, removing goal scoring improves MBPP pass@1 from $0.875$ to $1.000$, but reduces HotpotQA F1 from $0.340$ to $0.295$ and increases cumulative QA time from 21.76~s to 43.72~s. Similarly, removing topology lowers quality on both workloads, while removing resource routing sharply reduces MBPP performance despite a modest increase in HotpotQA F1. The semantic-only policy uses the fewest MBPP nodes but performs weakly on both quality metrics. Overall, the full AGAO configuration provides the most consistent balance of coding accuracy, QA quality, and execution efficiency across the two suites.

\section{Discussion}

\subsection{Why Graph Systems Need Attention}

The emergence of graph-based agent systems represents an important transition
from isolated reasoning loops toward structured collaboration. Graphs provide
explicit representations of task dependencies, agent interactions, and
execution workflows.

However, structure alone does not guarantee intelligent computation
allocation. In large-scale agent graphs, the number of available execution
paths can grow rapidly, while only a small subset of agents may contribute
significantly to the current objective.

This phenomenon resembles the evolution of neural architectures before the
introduction of attention mechanisms. Recurrent and convolutional models were
able to process sequential and spatial information, but they lacked an
effective mechanism to dynamically prioritize important information.

Similarly, graph-based agent systems provide connectivity, but require an
additional mechanism to determine computational focus.

Therefore, we argue that future agent systems require both:

\begin{equation}
\textbf{Graph}
+
\textbf{Attention}
=
\textbf{Adaptive Intelligence}
\end{equation}

where graph structures define possible interactions and attention mechanisms
determine which interactions deserve computational resources.

\subsection{Attention Engineering as a New Abstraction Layer}

We view Attention Engineering as an emerging abstraction layer for future
agentic systems.

Traditional agent frameworks focus on constructing workflows and connecting
capabilities. However, as the number of agents increases, manually designed
execution policies become increasingly difficult to maintain.

Attention Engineering introduces a dynamic control layer that enables systems
to automatically decide:

\begin{itemize}

\item which agents should participate;

\item which reasoning paths should be prioritized;

\item how much computation each agent should receive.

\end{itemize}

This perspective shifts agent design from
\textit{"building larger graphs"}
toward
\textit{"building adaptive computational focus mechanisms"}.

\subsection{Relationship with Existing Paradigms}

Attention mechanisms have been widely studied in different computational
settings. However, Attention Engineering differs from previous approaches in
both abstraction level and optimization objective.

Transformer attention operates on token sequences and aims to improve
representation learning. Graph attention mechanisms focus on learning better
node representations. Mixture-of-Experts routing dynamically activates neural
experts to improve model scalability.

In contrast, AGAO operates at the workflow execution level. Its objective is
not to learn better representations, but to allocate computational focus
among heterogeneous agents embedded in a dynamic execution graph.

Therefore, Attention Engineering complements existing attention mechanisms by
extending attention from information processing toward computational
orchestration.

\subsection{Towards Attention-native Agent Systems}

We envision that future autonomous agent platforms will evolve beyond static
workflow execution engines into attention-native operating systems.

Such systems may continuously maintain an attention state over:

\begin{itemize}

\item available agents;

\item external tools;

\item knowledge sources;

\item computational resources.

\end{itemize}

Instead of explicitly programming every workflow transition, developers may
specify high-level objectives, while attention mechanisms dynamically
construct and optimize execution paths.

This evolution suggests a future paradigm:

\begin{equation}
\text{Loop Engineering}
\rightarrow
\text{Graph Engineering}
\rightarrow
\text{Attention Engineering}
\end{equation}

where each stage represents an increasing level of abstraction in autonomous
system design.

\section{Conclusion}

In this paper, we introduce
\textbf{Attention Engineering}, a new execution paradigm for large-scale
multi-agent graph systems.

While graph-based agent frameworks provide structured representations for
complex collaboration, they suffer from attention dilution when the number of
agents and execution paths increases. Existing workflow engines primarily
focus on graph construction and execution ordering, but lack mechanisms to
dynamically determine which agents deserve computational focus.

To address this challenge, we propose
\textbf{Adaptive Goal-aware Attention Orchestration (AGAO)}, which introduces
an attention-driven execution layer between user objectives and agent graph
execution.

AGAO integrates three complementary attention mechanisms:

\begin{itemize}

\item
\textbf{Goal-aware Attention}, which measures semantic alignment between
user objectives and agent capabilities;

\item
\textbf{Topology-aware Attention}, which incorporates graph dependencies and
structural importance;

\item
\textbf{Resource-aware Attention}, which converts attention scores into
adaptive computation allocation policies.

\end{itemize}

Furthermore, we introduce adaptive graph routing, enabling attention states
to evolve according to intermediate execution feedback.

Extensive experiments on the proposed MAG-Focus benchmark demonstrate that
AGAO improves task performance while significantly reducing unnecessary agent
execution and computational cost.

Our work suggests a broader perspective:
as agent systems evolve from loop-based execution toward graph-based
collaboration, future autonomous systems will require mechanisms not only to
organize computation, but also to intelligently focus computation.

% \bibliographystyle{alpha}
% \bibliography{sample}

% ------------------------------------------------
% References
% ------------------------------------------------

\bibliographystyle{plainnat}

\bibliography{reference}

\end{document}